\documentclass[letterpaper, 10 pt, conference]{ieeeconf}  % Comment this line out if you need a4paper

\usepackage{graphicx}
\usepackage{subfig}
\usepackage{booktabs}
\usepackage{cite}
\usepackage{amsmath}
\usepackage{amssymb}
\usepackage{algorithm}
\usepackage{algorithmic}

\IEEEoverridecommandlockouts                              % This command is only needed if 
\usepackage{mathtools}
\usepackage{makecell}

\title{\LARGE \bf
Average-Reward Maximum Entropy Reinforcement Learning for Global Policy in Double Pendulum Tasks
}

\author{
Jean Seong Bjorn Choe$^{1}$,
Bumkyu Choi$^{2}$ and Jong-kook Kim$^{3}$
\thanks{$^{1}$School of Electrical Engineering, Korea University, Seoul, South Korea
    {\tt\small garangg@korea.ac.kr}}
\thanks{$^{2}$Independent Researcher
    {\tt\small qjarb3411@korea.ac.kr}}
\thanks{$^{3}$School of Electrical Engineering, Korea University, Seoul, South Korea
    {\tt\small jongkook@korea.ac.kr}}
}

\begin{document}

\maketitle
\thispagestyle{empty}
\pagestyle{empty}

%%%%%%%%%%%%%%%%%%%%%%%%%%%%%%%%%%%%%%%%%%%%%%%%%%%%%%%%%%%%%%%%%%%%%%%%%%%%%%%%
\begin{abstract}
This report presents our reinforcement learning-based approach for the swing-up and stabilisation tasks of the acrobot and pendubot, tailored specifcially to the updated guidelines of the 3rd AI Olympics at ICRA 2025. Building upon our previously developed Average-Reward Entropy Advantage Policy Optimization (AR-EAPO) algorithm, we refined our solution to effectively address the new competition scenarios and evaluation metrics. Extensive simulations validate that our controller robustly manages these revised tasks, demonstrating adaptability and effectiveness within the updated framework. 

% This report presents a solution for the swing-up and stabilisation tasks of the acrobot and the pendubot, developed for the AI Olympics competition at ICRA 2025. Our approach employs the Average-Reward Entropy Advantage Policy Optimization (AR-EAPO), a model-free reinforcement learning (RL) algorithm that combines average-reward RL and maximum entropy RL. 

\end{abstract}

%%%%%%%%%%%%%%%%%%%%%%%%%%%%%%%%%%%%%%%%%%%%%%%%%%%%%%%%%%%%%%%%%%%%%%%%%%%%%%%%
\section{Introduction}

The 3rd AI Olympics\footnote[1]{https://ai-olympics.dfki-bremen.de/} at ICRA 2025 continues the established tradition of evaluating advanced robotic control methods through standardised and challenging tasks involving underactuated double pendulum systems \cite{wiebe2022realaigym, wiebe2023open}. 
Building upon prior competitions at IJCAI 2023 \cite{ijcai2024p1043} and IROS 2024 \cite{wiebe2025reinforcement}, the current edition places particular emphasis on global policy robustness, requiring solutions for reliable swing-up stabilisation tasks from arbitrary initial configurations under significantly increased external disturbances.

The competition maintains its use of two different configurations: the acrobot, characterised by an inactive shoulder joint, and the pendubot, with an inactive elbow joint. 
Successfully addressing the newly defined criteria involves overcoming notable reinforcement learning (RL) challenges, such as ensuring consistent performance across diverse initial states and managing larger disturbances without extensive reward engineering or explicit system modelling.

Responding to these challenges, we leverage our previously developed Average-Reward Entropy Advantage Policy Optimisation (AR-EAPO) approach \cite{bjorn2024average, choe2025}, a model-free reinforcement learning method. 

In the previous competition at IROS 2024, AR-EAPO demonstrated strong performance, winning the pendubot category and finishing runner-up in the acrobot category. This report outlines our updated methodology and demonstrates its effectiveness through comprehensive simulation experiments, explicitly focusing on meeting the robustness and adaptability standards set by the updated competition guidelines.

\section{Backgrounds}
\subsection{Average-Reward MaxEnt Reinforcement Learning}
This work considers unichain Markov decision processes (MDP) $\langle\mathcal{S}, \mathcal{A}, r, p\rangle$, where $\mathcal{S}$ is the state space, $\mathcal{A}$ denotes the action space, $r$ is the reward function, and $p$ is the transition kernel. We define average-reward maximum entropy (MaxEnt) objective and a stationary policy $\pi: \mathcal{S} \rightarrow \Delta(\mathcal{A})$, where $\Delta(\mathcal{A})$ denotes the set of probability distributions over $\mathcal{A}$, 
\begin{align}
\label{eq:soft-gain}
\rho^\pi_\tau \coloneqq 
   \lim_{T\rightarrow\infty}\frac{1}{T}\,
   \mathop{\mathbb{E}}\limits_{\substack{S_0 = s, \\ A_t \sim \pi}}
   \left[\sum_{t=0}^{T-1} \Bigl(R_t \;-\; \tau \log\pi(A_t \mid S_t)\Bigr)\right],
\end{align}
where $S_t$ is the state at time $t$ and $A_t$ is the action at time $t$, $R_t=r(S_t,A_t)$ is the reward signal at time $t$, $s$ is an arbitrary starting state, and $\tau$ is the temperature parameter adjusting the entropy regularisation. This objective, called gain or average reward, aims to maximise the long-term average reward instead of the sum of discounted rewards. It's especially valuable for continuing tasks with no defined endpoint, as it offers a more intuitive method for evaluating policies without requiring artificial discount factors.

Since the gain is independent of the starting state $s$ in unichain MDPs, it is convenient to define a function to quantify the relative advantage of starting in a particular state compared to the average performance. We define the bias value function $v(s)$ for state $s\in\mathcal{S}$ that satisfies the following Bellman equation:
\begin{align*}
% \label{eq:vf}
    v_\tau^\pi(s) + \rho^\pi_\tau = \mathop{\mathbb{E}}\limits_{a\sim\pi}\left[r(s,a)-\tau\log\pi(a|s) + \sum_{s'\in\mathcal{S}}p^{s'}_{s,a}v_\tau^\pi(s')\right],
\end{align*}
where we denote $p^{s'}_{s,a}$ as the transition probability to state $s'$ when performing action $a$ from state $s$ .

\subsection{Average-Reward Entropy Advantage Policy Optimisation}
AR-EAPO \cite{choe2025} is an actor-critic algorithm to optimise the average-reward MaxEnt objective (\ref{eq:soft-gain}). Based on the concept of the bias value function, the method estimates the advantage function $a^\pi: \mathcal{S}\times\mathcal{A} \mapsto \mathbb{R}$, which measures how much better an action is compared to the average performance in a given state:
\begin{align*}
a^\pi(s,a) &\coloneqq r_\tau(s,a) -\rho^\pi_\tau + \mathbb{E}_{s'\sim p}[v^\pi_\tau(s')] - v^\pi_\tau(s), \\
r_\tau &\coloneqq r(s,a) - \tau\log\pi(a|s).
\end{align*}
% trajectories sampled from the policy $\pi$ and 
AR-EAPO constructs the advantage function by decomposing the bias value function into the reward and the entropy parts and applying different amounts of variance reduction \cite{schulman2015high} for each. Then it estimates the advantage of sampled states and actions to update the gain and the parameterised policy $\pi_\theta$ using the PPO objective \cite{schulman2017proximal}.

\section{Proposed Method}
Our approach to achieving a global controller focuses on modifying the MDP design rather than the policy optimisation. Specifically, we adjust the initial state distribution and the random truncation probability \cite{choe2025} to simulate the global swing-up environment without explicitly incorporating the evaluation protocol's random reset process. 

We consider 4-D state space $\mathcal{S}\subset\mathbb{R}^4$ with $s=[q_1,q_2,\dot q_1,\dot q_2]^\intercal\in\mathcal{S}$ where $q_i$ represents the wrapped angle of joint $i$ and $\dot q_i$ is the normalised angular velocity of joint $i$. We model the initial state distribution $\mu_0$ as a Gaussian distribution centred on the bottom state $s_0$ with a large variance $\sigma=6.0$:
\begin{align*}
    s_0'\sim\mu_0=\mathcal{N}(s_0,\sigma),
\end{align*}
where $s_0=[0, 0, 0, 0]^\intercal$.

For the action space, we use a 1-dimensional continuous action space as each tasks allows the agent to control only one joint.
% The output of the policy network corresponds directly to the normalised torque input.
Therefore, the parametrised policy network $f_\theta(s):\mathcal{S}\mapsto \mathbb{R}\times\mathbb{R}^+$ maps a state vector $s$ to the mean $\mu_\theta(s)$ and the standard deviation $\sigma_\theta(s)$ of a squashed Gaussian distribution, i.e.,
\begin{align*}
    \mu_\theta(s),\sigma_\theta(s) &= f_\theta(s), \\
    a\sim\pi_\theta(\cdot|s)&=\tanh(\mathcal{N}(\mu_\theta(s),\sigma_\theta(s))),
\end{align*}
where $a$ corresponds to a normalised torque input.

In contrast to the previous competition where excessive velocity required penalisation for efficient swing-ups, we employ a more relaxed quadratic cost and omit the torque penalty to allow faster goal state achievement:
\begin{align*}
    r(s,a) = -\alpha[(s -g)^\intercal Q(s-g)],
\end{align*}
where $g=[\pi,0,0,0]^\intercal$ is the goal state with scaling factor $\alpha=0.001$ and cost matrix $Q=\text{diag}([100, 100, 4, 2])$. 

We introduce a large random truncation probability $p_\text{trunc}=0.005$ to induce bias toward faster swing-up during training. Specifically, $p_\text{trunc}$ effectively limits the temporal horizon over which the average reward is calculated to approximately $1/(1-p_\text{trunc})$ steps in expectation. Combined with the large reset variance, this causes the value estimate to rely more heavily on short-range bootstrapping with immediate costs during the swing-up process. As a result, policies that achieve faster swing-ups are favoured, since the algorithm has less opportunity to amortise the initial swing-up cost over a longer trajectory when computing the average reward.

\section{Results}
The performance metric\footnote[1]{https://ai-olympics.dfki-bremen.de/} for the current competition has been simplified, measuring the total time the system remains within the designated goal region at the upright (swing-up) position during each 60-second trial. 
However, particular emphasis is placed on robust performance under significantly increased external disturbances. 
Unlike previous editions, the evaluation explicitly prioritises the controller’s resilience against external noise and perturbations. 

% We directly incorporated these updated evaluation criteria into our training phase, selecting the best-performing policy based on its robustness and ability to handle external disturbances.

% \textbf{During training phase, we directly incorporated these updated evaluation criteria by periodically resetting the system state to random joint angels and velocities. Specifically, at random times (up to 15 per 60-second episode), the controller was momentarily overridden for a short duration (max. 1s), forcing the policy network to recover robustly from these disturbances.
% Accordingly, we selected the best-performing policy based on its ability to reliably maintain or quickly restore the upright position under such conditions.}

% simulating external disturbances. Specifically, we introduced random perturbations to the state vector, representing it intermittently to random positions and velocities sampled uniformly from the entire states space. These disturbances were applied approximately every 3 seconds during each 60-second episode, ensuring a PID ,,,
% Additionally, we selected the best-performing policy based on ints robustness and ability to quickly and reliably return the system to the upright position following these disturbances.

We employed a policy network with two hidden layers of size [256, 256] and a value network with two hidden layers of size [512, 512], both using ReLU activations. 
Each policy underwent multiple training runs, spanning 30 million frames each, with periodic performance evaluations using the update criterion.
The best-performing policy from these runs was selected for reporting. 
This process took approximately 100 minutes on a single system equipped with an AMD Ryzen 9 5900X CPU and NVIDIA RTX 3080 GPU. We observed initial convergence at around 20 million frames.
 
Table \ref{table:hyperparmeters} summarizes the hyperparameters utilised for AR-EAPO training, noting a noise variance of 4.0 to better handle stronger noise conditions.

\begin{table}[h]
    \centering
    \caption{Hyperparmeters used for training.}
    \begin{tabular}{lclc}
        % \toprule
        \textbf{Hyperparmeters} & \textbf{Value} & \textbf{Hyperparameters (cont.)} & \textbf{Value} \\
        \midrule
        temperature $\tau$ & 1.5  & log std. init. & 0.5 \\
        reward GAE $\lambda$ & 0.8  & adv. minibatch norm. & True \\
        entropy GAE $\lambda_e$ & 0.6  &  max grad. norm. & 10.0 \\
        PPO clip range $\epsilon$ & 0.05 & num. envs. & 64 \\
        gain step size $\eta$ & 0.01 & num. rollout steps & 128 \\
        learning rate & $5\mathrm{e}{-4}$ & num. epochs & 6 \\
        $p_\text{trunc}$ & $5\mathrm{e}{-3}$ & batch size & 1024 \\
        EAPO $c_2$& 0.5 & vf. coef. & 0.25 \\
        reset noise variance & 4.0 \\
        \bottomrule
    \end{tabular}
    \label{table:hyperparmeters}
\end{table}

\begin{table}[h]
    \centering
    \caption{RealAI Score of the Acrobot for 5 different random seeds (35, 177, 1670, 334, 15793).}

    \resizebox{0.4866\textwidth}{!}{
    \begin{tabular}{lcccc}
        \toprule
        \textbf{Trial \#} & \textbf{AR-EAPO (Ours)} & \textbf{AR-EAPO (v2)} & \textbf{MCPILCO} & \textbf{TVLQR} \\
        \midrule
        Trial\_1 & 0.702 & 0.440 & 0.087 & 0.005 \\
        Trial\_2 & 0.642 & 0.345 & 0.0 (E) & 0.008 \\
        Trial\_3 & 0.680 & 0.332 & 0.0 (E) & 0.013 \\
        Trial\_4 & 0.670 & 0.349 & 0.092 & 0.005 \\
        Trial\_5 & 0.649 & 0.371 & 0.0 (E) & 0.008 \\
        \textbf{Avg Score} & \textbf{0.667} & \textbf{0.367} & \textbf{0.037} & \textbf{0.008} \\
    \end{tabular}
    }

    \vspace{0.3em}

    \resizebox{0.4866\textwidth}{!}{
    \begin{tabular}{lcccc}
        \toprule
        \textbf{} & \textbf{ILQR MPC} & \textbf{ILQR Riccati} & \textbf{EVOL SAC} & \textbf{HISTORY-SAC} \\
        \midrule
        Trial\_1 & 0.000 & 0.036 & 0.319 & 0.122 \\
        Trial\_2 & 0.019 & 0.019 & 0.399 & 0.132 \\
        Trial\_3 & 0.012 & 0.028 & 0.309 & 0.152 \\
        Trial\_4 & 0.000 & 0.026 & 0.376 & 0.165 \\
        Trial\_5 & 0.000 & 0.018 & 0.278 & 0.182 \\
        \textbf{Avg Score} & \textbf{0.006} & \textbf{0.025} & \textbf{0.336} & \textbf{0.151} \\
        \bottomrule
    \end{tabular}
    }

    \label{table:performance_scores_acrobot}
\end{table}

% \begin{table}[!h]
%     \centering
%     \caption{RealAI Score of the Pendubot for 5 different random seeds (6362, 1709, 49219, 83, 558).}
%     \resizebox{0.4866\textwidth}{!}
%     {
%     \begin{tabular}{lccccccccc}
%         % \toprule
%         \textbf{Trial \#} & \textbf{AR-EAPO (Ours)} & \textbf{AR-EAPO (v2)} & \textbf{MCPILCO} & \textbf{TVLQR} &  \textbf{ILQR MPC} &  \textbf{ILQR Riccati} & \textbf{EVOL SAC} & \textbf{HISTORY-SAC} \\
%         \midrule
%          Trial\_1  & 0.793 & 0.587 & 0.096 & 0.124 & 0.125 & 0.021 & 0.369 & 0.369 \\
%          Trial\_2  & 0.792 & 0.629 & 0.0(E) & 0.004 & 0.017 & 0.015 & 0.186 & 0.369 \\
%          Trial\_3  & 0.749 & 0.666 & 0.0(E) & 0.118 & 0.119 & 0.034 & 0.288 & 0.369 \\
%          Trial\_4  & 0.711 & 0.662 & 0.101 & 0.123 & 0.116 & 0.025 & 0.233 & 0.369 \\
%          Trial\_5  & 0.714 & 0.69 & 0.093 & 0.007 & 0.021 & 0.021 & 0.408 & 0.369 \\
%         \textbf{Avg Score} & \textbf{0.752} & \textbf{0.647} & \textbf{0.06} & \textbf{0.075} & \textbf{0.080} & \textbf{0.023} & \textbf{0.297} & \textbf{0.297} \\
%         \bottomrule
%     \end{tabular}
%     }
%     \label{table:performance_scores_pendubot}
% \end{table}

\begin{table}[h]
    \centering
    \caption{RealAI Score of the Pendubot for 5 different random seeds (6362, 1709, 49219, 83, 558).}

    \resizebox{0.4866\textwidth}{!}{
    \begin{tabular}{lcccc}
        \toprule
        \textbf{Trial \#} & \textbf{AR-EAPO (Ours)} & \textbf{AR-EAPO (v2)} & \textbf{MCPILCO} & \textbf{TVLQR} \\
        \midrule
        Trial\_1 & 0.793 & 0.587 & 0.096 & 0.124 \\
        Trial\_2 & 0.792 & 0.629 & 0.0 (E) & 0.004 \\
        Trial\_3 & 0.749 & 0.666 & 0.0 (E) & 0.118 \\
        Trial\_4 & 0.711 & 0.662 & 0.101 & 0.123 \\
        Trial\_5 & 0.714 & 0.690 & 0.093 & 0.007 \\
        \textbf{Avg Score} & \textbf{0.752} & \textbf{0.647} & \textbf{0.060} & \textbf{0.075} \\
    \end{tabular}
    }

    \vspace{0.3em}

    \resizebox{0.4866\textwidth}{!}{
    \begin{tabular}{lcccc}
        \toprule
        \textbf{} & \textbf{ILQR MPC} & \textbf{ILQR Riccati} & \textbf{EVOL SAC} & \textbf{HISTORY-SAC} \\
        \midrule
        Trial\_1 & 0.125 & 0.021 & 0.369 & 0.396 \\
        Trial\_2 & 0.017 & 0.015 & 0.186 & 0.316 \\
        Trial\_3 & 0.119 & 0.034 & 0.288 & 0.361 \\
        Trial\_4 & 0.116 & 0.025 & 0.233 & 0.37 \\
        Trial\_5 & 0.021 & 0.021 & 0.408 & 0.318 \\
        \textbf{Avg Score} & \textbf{0.080} & \textbf{0.023} & \textbf{0.297} & \textbf{0.352} \\
        \bottomrule
    \end{tabular}
    }

    \label{table:performance_scores_pendubot}
\end{table}

Due to the stochastic nature of the disturbances, we evaluated our AR-EAPO controllers across five different random seeds. To enable a fair comparison, we additionally tested previous benchmark controllers and participant-submitted controllers \cite{wiebe2025reinforcement, turcato2024learning, cali2024ai, lukas2024velocity}, including the previous version of our own controller (AR-EAPO v2)\cite{bjorn2024average}, under the same seeds. 
Those controllers were tested in their original form, without modifications to reflect updated competition guidelines such as the revised maximum velocity limit. The notation (E) in the tables indicates episodes with numerical instability or NaN errors occurred during executions.
The results, summarised in Tables \ref{table:performance_scores_acrobot} and \ref{table:performance_scores_pendubot}, demonstrate that AR-EAPO consistently achieves superior and stable performance for both the acrobot and pendubot scenarios.

% Fig.\ref{swing-up_acrobot} and Fig.\ref{swing-up_penubot} illustrate the time-series plots of position, velocity, and torque for the acrobot and pendubot, respectively. 

\begin{figure}[!h]
      \centering
      \includegraphics[width=0.486\textwidth, height=0.5\linewidth]{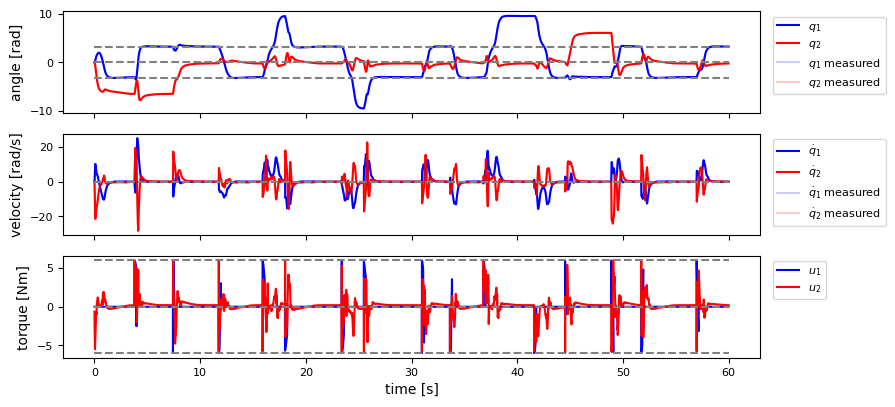}
      \caption{Swing-up trajectory with AR-EAPO on the acrobot (the best-performing seed out of the five random seeds we evaluated).}
      \label{swing-up_acrobot}
   \end{figure}

\begin{figure}[!h]
      \centering
      \includegraphics[width=0.486\textwidth, height=0.5\linewidth]{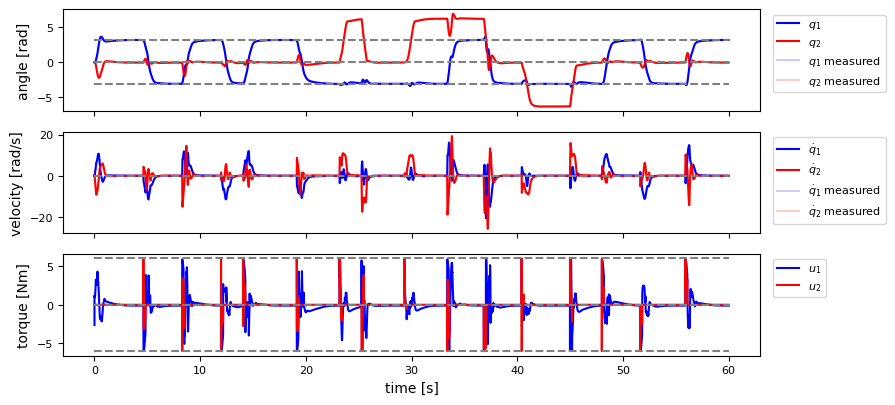}
      \caption{Swing-up trajectory with AR-EAPO on the pendubot (the best-performing seed out of the five random seeds we evaluated).}
      \label{swing-up_penubot}
   \end{figure}

Fig.\ref{swing-up_acrobot} and Fig.\ref{swing-up_penubot} illustrate the time-series plots of angles $(q_1, q_2)$, angular velocities $(\dot{q_1},\dot{q_2})$, and torque $(\tau)$ for the acrobot and pendubot, respectively. 
% The variables $q_1, q_2$ denote the joint angels, and $(\dot{q_1},\dot{q_2})$ represents the angular velocities of the two degrees of freedom (DOFs).
The dashed horizontal lines represent reference values: joint angles at $\pm \pi$, angular velocities at $0$, and torque limits at $\pm \tau_{limit}$.
These plots clearly demonstrate the controller’s ability to recover and stabilise following periodic random disturbances.

\section{conclusion}

In this report, we presented our reinforcement learning-based solution for the swing-up and stabilisation tasks of the acrobot and pendubot, building upon our previous work using the Average-Reward Entropy Advantage Policy Opimization (AR-EAPO), a model-free RL algorithm. We successfully adapted our approach to address new competition conditions and significantly increased disturbances introduced in the 3rd AI Olympics. The Simulation results confirmed the effectiveness and adaptability of extended AR-EAPO method.

%%%%%%%%%%%%%%%%%%%%%%%%%%%%%%%%%%%%%%%%%%%%%%%%%%%%%%%%%%%%%%%%%%%%%%%%%%%%%%%%

%%%%%%%%%%%%%%%%%%%%%%%%%%%%%%%%%%%%%%%%%%%%%%%%%%%%%%%%%%%%%%%%%%%%%%%%%%%%%%%%



% \section*{ACKNOWLEDGMENT}

% The preferred spelling of the word ÒacknowledgmentÓ in America is without an ÒeÓ after the ÒgÓ. Avoid the stilted expression, ÒOne of us (R. B. G.) thanks . . .Ó  Instead, try ÒR. B. G. thanksÓ. Put sponsor acknowledgments in the unnumbered footnote on the first page.

%%%%%%%%%%%%%%%%%%%%%%%%%%%%%%%%%%%%%%%%%%%%%%%%%%%%%%%%%%%%%%%%%%%%%%%%%%%%%%%%

\bibliographystyle{IEEEtran}
\bibliography{ref.bib}

\begin{thebibliography}{10}
\providecommand{\url}[1]{#1}
\csname url@samestyle\endcsname
\providecommand{\newblock}{\relax}
\providecommand{\bibinfo}[2]{#2}
\providecommand{\BIBentrySTDinterwordspacing}{\spaceskip=0pt\relax}
\providecommand{\BIBentryALTinterwordstretchfactor}{4}
\providecommand{\BIBentryALTinterwordspacing}{\spaceskip=\fontdimen2\font plus
\BIBentryALTinterwordstretchfactor\fontdimen3\font minus \fontdimen4\font\relax}
\providecommand{\BIBforeignlanguage}[2]{{%
\expandafter\ifx\csname l@#1\endcsname\relax
\typeout{** WARNING: IEEEtran.bst: No hyphenation pattern has been}%
\typeout{** loaded for the language `#1'. Using the pattern for}%
\typeout{** the default language instead.}%
\else
\language=\csname l@#1\endcsname
\fi
#2}}
\providecommand{\BIBdecl}{\relax}
\BIBdecl

\bibitem{wiebe2022realaigym}
F.~Wiebe, S.~Vyas, L.~Maywald, S.~Kumar, and F.~Kirchner, ``Realaigym: Education and research platform for studying athletic intelligence,'' in \emph{Proceedings of Robotics Science and Systems Workshop Mind the Gap: Opportunities and Challenges in the Transition Between Research and Industry, New York}, 2022.

\bibitem{wiebe2023open}
F.~Wiebe, S.~Kumar, L.~J. Shala, S.~Vyas, M.~Javadi, and F.~Kirchner, ``Open source dual-purpose acrobot and pendubot platform: Benchmarking control algorithms for underactuated robotics,'' \emph{IEEE Robotics \& Automation Magazine}, vol.~31, no.~2, pp. 113--124, 2023.

\bibitem{ijcai2024p1043}
\BIBentryALTinterwordspacing
F.~Wiebe, N.~Turcato, A.~Dalla~Libera, C.~Zhang, T.~Vincent, S.~Vyas, G.~Giacomuzzo, R.~Carli, D.~Romeres, A.~Sathuluri, M.~Zimmermann, B.~Belousov, J.~Peters, F.~Kirchner, and S.~Kumar, ``Reinforcement learning for athletic intelligence: Lessons from the 1st “ai olympics with realaigym” competition,'' in \emph{Proceedings of the Thirty-Third International Joint Conference on Artificial Intelligence, {IJCAI-24}}, K.~Larson, Ed.\hskip 1em plus 0.5em minus 0.4em\relax International Joint Conferences on Artificial Intelligence Organization, 8 2024, pp. 8833--8837, demo Track. [Online]. Available: \url{https://doi.org/10.24963/ijcai.2024/1043}
\BIBentrySTDinterwordspacing

\bibitem{wiebe2025reinforcement}
F.~Wiebe, N.~Turcato, A.~D. Libera, J.~S.~B. Choe, B.~Choi, T.~L. Faust, H.~Maraqten, E.~Aghadavoodi, M.~Cali, A.~Sinigaglia \emph{et~al.}, ``Reinforcement learning for robust athletic intelligence: Lessons from the 2nd'ai olympics with realaigym'competition,'' \emph{arXiv preprint arXiv:2503.15290}, 2025.

\bibitem{bjorn2024average}
J.~S. Bjorn~Choe, B.~Choi, and J.-k. Kim, ``Average-reward maximum entropy reinforcement learning for underactuated double pendulum tasks,'' \emph{arXiv e-prints}, pp. arXiv--2409, 2024.

\bibitem{choe2025}
J.~S.~B. Choe, B.~Choi, and J.-k. Kim, ``Simplifying reward design in complex robotics:\\ average-reward maximum entropy reinforcement learning,'' in \emph{2025 IEEE International Conference on Robotics and Automation (ICRA)}.\hskip 1em plus 0.5em minus 0.4em\relax IEEE, 2025.

\bibitem{schulman2015high}
J.~Schulman, P.~Moritz, S.~Levine, M.~Jordan, and P.~Abbeel, ``High-dimensional continuous control using generalized advantage estimation,'' \emph{arXiv preprint arXiv:1506.02438}, 2015.

\bibitem{schulman2017proximal}
J.~Schulman, F.~Wolski, P.~Dhariwal, A.~Radford, and O.~Klimov, ``Proximal policy optimization algorithms,'' \emph{arXiv preprint arXiv:1707.06347}, 2017.

\bibitem{turcato2024learning}
N.~Turcato, A.~D. Libera, G.~Giacomuzzo, R.~Carli, and D.~Romeres, ``Learning control of underactuated double pendulum with model-based reinforcement learning,'' \emph{arXiv preprint arXiv:2409.05811}, 2024.

\bibitem{cali2024ai}
M.~Cal{\`\i}, A.~Sinigaglia, N.~Turcato, R.~Carli, and G.~A. Susto, ``Ai olympics challenge with evolutionary soft actor critic,'' \emph{arXiv preprint arXiv:2409.01104}, 2024.

\bibitem{lukas2024velocity}
T.~Lukas~Faust, H.~Maraqten, E.~Aghadavoodi, B.~Belousov, and J.~Peters, ``Velocity-history-based soft actor-critic tackling iros'24 competition" ai olympics with realaigym",'' \emph{arXiv e-prints}, pp. arXiv--2410, 2024.

\end{thebibliography}

\end{document}